\documentclass[10pt,twocolumn,letterpaper]{article}

\usepackage{cvpr}
\usepackage{times}
\usepackage{epsfig}
\usepackage{graphicx}
\usepackage{caption}
\usepackage{subcaption}
\usepackage{amsmath}
\usepackage{amssymb}
\usepackage{adjustbox}
\usepackage{graphics}
\usepackage{booktabs}
\usepackage{multirow}
\usepackage{makecell}

\usepackage[british]{babel}
\usepackage{hhline}
\usepackage[breaklinks=true,bookmarks=false]{hyperref}

\cvprfinalcopy % *** Uncomment this line for the final submission

\def\cvprPaperID{31} % *** Enter the CVPR Paper ID here

\ifcvprfinal\fi
\begin{document}

%%%%%%%%% TITLE
\title{Climate Adaptation: Reliably Predicting from Imbalanced Satellite Data}

\author{
Ruchit Rawal\thanks{Equal contribution.}\\
Netaji Subhas University of Technology,\\
New Delhi, India.\\
{\tt\small ruchitr.ec.17@nsit.net.in}
\and
Prabhu Pradhan\footnotemark[1] \thanks{Part of the work was done at GCDSL, Aerospace Engineering, Indian Institute of Science (IISc Bangalore) \scriptsize{\url{https://bit.ly/gcdsl_iisc}}}\\
Max Planck Institute for Intelligent Systems,\\
T\"ubingen, Germany.\\
{\tt\small prabhu.pradhan@tuebingen.mpg.de}
}
\maketitle
%\thispagestyle{empty}
%%%%%%%%% ABSTRACT
\begin{abstract}
The utility of aerial imagery (Satellite, Drones) has become an invaluable information source for cross-disciplinary applications, especially for crisis management. Most of the mapping and tracking efforts are manual which is resource-intensive and often lead to delivery delays. Deep Learning methods have boosted the capacity of relief efforts via recognition, detection, and are now being used for non-trivial applications. However the data commonly available is highly imbalanced (similar to other real-life applications) which severely hampers the neural network's capabilities, this reduces robustness and trust. We give an overview on different kinds of techniques being used for handling such extreme settings and present solutions aimed at maximizing performance on minority classes using a diverse set of methods (ranging from architectural tuning to augmentation) which as a combination generalizes for all minority classes. We hope to amplify cross-disciplinary efforts by enhancing model reliability.
\end{abstract}
\thispagestyle{empty}
\section{Introduction}
The last decade has witnessed tremendous growth both in computational power and scientific methods for pattern recognition and data science. Machine Learning is a tool driving many technologies across diverse sectors. However the fuel that drives this growth is data, and as is with every fuel it's not directly usable. A critical problem is class imbalance, both in supervised and unsupervised form of learning algorithms.
A dataset can be treated as imbalanced if there is a noticeable mismatch between the target variable and other values. 
For example, medical-diagnostics data is conventionally biased towards the negative class (healthy samples are more numerous than the infected ones). 
Other examples include fraud detection, natural language processing, visual recognition, astronomy, etc. Experimentally, high instability in performance has been observed in vanilla models when tested on imbalanced datasets  \cite{Buda2018ASS}.

Commonly, deep-net models are built to maximize predictive accuracy (ex. classification) but this metric is uneventful for the cases with limited labels, extreme classification etc.  \cite{Lipton19}. This happens because the trained classifier focuses only on the most-numerous class (since it has a higher proportion) while remaining below-par on minority classes. This may prove catastrophic in critical use cases like medical diagnostics and self-driving cars where the rare instances are of utmost importance.

Our use case consists of satellite imagery of African region which is labelled to help automate the process of predicting drought, cattle sustenance etc. via estimating the quality of forage  \cite{droughtdataWandB}. Usually, non-profit organizations cannot employ a dedicated team of ML engineers/researchers or clusters of GPUs  \cite{Ferreira2019AnAN}, models that can perform robustly and reliably at low requirements can be pragmatically utilized by domain-experts and local administration \cite{Pradhan2019SmarterPF} for easy deployment.

Presently, researchers tend to tackle the imbalance issues (either at input or intermediate pipeline) in its narrow context with domain-specific solutions. We present drawn-out insights on several techniques to mitigate data-imbalance problems.
The contributions of this paper are:
\begin{itemize}
\item We use a deep generative model for synthetic data augmentation of multi-spectral images. To the best of our knowledge, this specific area is still unexplored.\\ We also show that certain spectral bands are better for particular tasks (here, vegetation area analysis).
\item We show that a combination of Cyclic Learning Rate (CLR)  \cite{CyclicalLR} + Stochastic Weight Averaging (SWA)  \cite{Izmailov2018AveragingWL} is suitable for extreme imbalance scenarios.
\item We further cement the compatibility of LDAM: Label-distribution aware loss-function  \cite{LDAM}, which works better than crude re-sampling and can be further improved by using class-balanced loss  \cite{Cui2019ClassBalancedLB}.
\end{itemize}

The rest of this paper is as follows: Section \ref{dataset} introduces the dataset. Section \ref{NN} provides details on our modifications to the base neural-net model. It also includes subsections on Loss Function (\ref{loss}) which gives an overview of sampling functions,  Cyclic Learning Rate (\ref{clr}) which is a popular training routine, and Stochastic Weight Averaging (\ref{swavg}) as a powerful regularizer for handling data-imbalance issues. Section \ref{msbands} provides details on multi-spectral imagery from the lens of machine learning. Section \ref{gans} presents our experiments with synthetic data augmentation followed by our overall results. Finally we conclude the paper in Section \ref{end} with a short discussion on performance metrics.

We use intra-class variance (ICV), Balanced accuracy  \cite{Brodersen2010TheBA} and Recall as performance metrics (definitions in Sec. \ref{metric}).

We present all our observations in two graph plots -\\\autoref{fig1} (ValAcc vs ICV) and \autoref{fig2} (BalAcc vs ICV).

Our codebase- \scriptsize{\url{https://github.com/JARVVVIS/drought}}. \normalsize

\subsection{Performance Metrics}
\label{metric}
As we see in  \cite{Lipton19, Sokolova2009ASA, RoccaTDS, BoyleTDS}, accuracy is not the best metric to evaluate imbalanced datasets, as it can be very misleading. Metrics that provide better insights  \cite{Sajjadi2018} include:
\begin{itemize}
    \item \textbf{Recall}: Recall portrays the fraction of true positives could be detected correctly, It is defined as $True Positive/{(True Positive + False Negative)}$, Thus a low recall signifies a high number of false negatives which is undesirable in a real-world setting.
    
    \item \textbf{Balanced Accuracy} (BalAcc): The arithmetic mean of the TPR (True Positive Rate) and TNR (True Negative Rate). Thus if the model is exploiting the class-imbalance problem itself to increase the vanilla accuracy, the balanced accuracy will drop significantly and reflect the poor performance.
    \item \textbf{Intra-Class Variance} (ICV) :   ${\sum_{i=0}^{4} {(acc- class_{i})}^2}$ \\(where $acc$ denotes the validation accuracy and $class_i$ denotes the accuracy of the $i^{th}$ class for a given experiment). The aim is expose the models which have high category-variance (per-class accuracies) owing to overfitting on the frequent class (in comparison to robust models i.e. low variance). \cite{Em2017IncorporatingIV} highlights importance of ICV for visual recognition tasks.
\end{itemize}
\autoref{MetricsTable} presents a comparative list of our final results as per the aforementioned benchmarks.

\section{Data-set} \label{dataset}
The expert-labelled, multi-spectral satellite \href{https://developers.google.com/earth-engine/datasets/catalog/LANDSAT_LC08_C01_T1_RT}{(LANDSAT) data}  \cite{droughtdataWandB}  was released as a bid to enhance drought detection pipelines. It  essentially consists of 100,000 images split into 86,317 training and 10,778 validation images, having a spatial resolution of 65x65 pixels over 10 spectrum bands. Each image is labeled by a human expert as- 'the number of cows the geographical location at the center of the image can support', serving as a measure of forage quality of the location and further as an indicator of whether the location is arid (drought-hit). 

The dataset is highly imbalanced (roughly ~60\% of the data gathered is of class 0, classes 1 and 2 have ~15\% each, and the remaining ~10\%  is class 3). The model can erroneously achieve 60\% accuracy just by predicting 0 every time. However, such high mis-classification is very problematic since these algorithms will be deployed in high-stake real-world settings. We would like to make dense predictions no matter the location of the pixel, since there is high amount of sparsity in the labels. Hence, we need to train a model that is satisfactorily robust to out-of-distribution (o.o.d) samples and generalizes well on all the inherent classes i.e. independent-\&-identically-distributed (i.i.d) samples. We focus on striking a pragmatic balance.

\section{Network Architectures} \label{NN}

Ever since winning the 2015 ILSVRC  \cite{imagenet} challenge ResNet  \cite{resnet} has inspired a family of deep convolutional neural networks. The skip connections in ResNet allow one to build deep networks (up to 1000 layers) while still keeping them optimizable, He et. al  \cite{resnet} demonstrated that even for fixed baseline architecture increase in depth almost always leads to increased accuracy.

While Scaling in depth  \cite{resnet} is the go-to method to boost a network's accuracy, other less popular scaling methods include scaling by width  \cite{Zagoruyko2016WideRN} and resolution  \cite{GPipe}. 
Tan et. al  \cite{Tan2019EfficientNetRM} in their work showed that while scaling (in width, depth, resolution) improves model accuracy, the accuracy saturates after a certain level. 
They argued that different scaling dimensions (height, width, resolution) are not independent and the key to successfully scale deep networks is in balancing scaling in different dimensions rather than scaling in one direction only. To harmonize the scaling in all dimensions they proposed a compound scaling method which utilized $\phi$ (compound scaling coefficient) to 
uniformly scales the network's depth, width and resolution.
\framebox[\columnwidth]{Depth : d =  $\alpha^\phi$,\hfill Width : w = $\beta^\phi$,\hfill Resolution : r = $\gamma^\phi$}
\framebox[\columnwidth]{
\indent s.t. $\alpha*\beta^2*\gamma^2\approx2$
\indent\indent $\alpha,\beta,\gamma\geq1$
}

However, scaling doesn't change the core layer operations making it imperative to have a solid baseline network for achieving desired outcomes, Tan et al.  \cite{Tan2019EfficientNetRM} leveraged Neural Architecture Search  \cite{NASZoph2016NeuralAS} to propose a new baseline "Efficient-Net" by optimizing for both accuracy and FLOPS.

\begin{table}[b]
\begin{adjustbox}{width=\columnwidth,center}
\begin{tabular}{cccccccccc}

    \toprule
    \multirow{2}{*}[-4pt]{\thead{Model}} &
    \multirow{2}{*}[-4pt]{\thead{Training Details}} &
    \multirow{2}{*}[-4pt]{\thead{Validation Acc}} &
    \multicolumn{4}{c}{\thead{Recall}}\\
    
    \cmidrule(lr){4-7}
    &&& 0 & 1 & 2 & 3\\
    \midrule
    ResNet-50 & \makecell{Learning Rate : 1e-3  \\ Loss : Cross-Entropy} & 0.7465 & 0.9102 & 0.4198 & 0.5511 & 0.5682 \tabularnewline
    \tabularnewline
    Efficient-Net B4 & \makecell{Learning Rate : 1e-3 \\ Loss : Cross-Entropy} & 0.7630 & 0.9199 & 0.4114 & 0.5739 & 0.6469 \tabularnewline
    
    \bottomrule
\end{tabular}
\end{adjustbox}
\caption{Comparison of Baseline Performances on Validation Set. Efficient-Net B4 attains higher accuracy and per-class Recall in comparison to ResNet-50.}
\label{NNarch}
\end{table}

In \autoref{NNarch} we give a baseline for ResNet-50 and Efficient-Net B4. We also apply standard data augmentation e.g. Random Horizontal-Flips, Random Vertical-Flips and Random Rotation after normalizing the data.

\subsection{Loss Function and Sampling} \label{loss}

\begin{table}[b]
\begin{adjustbox}{width=\columnwidth,center}
\begin{tabular}{cccccccccc}
    \toprule
    \multirow{2}{*}[-4pt]{\thead{Model}} &
    \multirow{2}{*}[-4pt]{\thead{Training Details}} &
    \multirow{2}{*}[-4pt]{\thead{Validation Acc}} &
    
    \multicolumn{4}{c}{\thead{Recall}}\\
    \cmidrule(lr){4-7}
    &&& 0 & 1 & 2 & 3\\
    \midrule
    ResNet-50 & \makecell{Learning Rate : 1e-3  \\ Sampler : $\propto 1/n_j$ \\ Loss : Cross-Entropy \\  }
    & 0.7184 & 0.8154 & 0.5818 & 0.5014 & 0.6880 \tabularnewline
    \tabularnewline
    ResNet-50 & \makecell{Learning Rate : CLR  \\ Sampler : $\propto 1/n_j $ \\ Loss : LDAM+DRW \\  }
    & 0.7022 & 0.7792 & 0.5719 & 0.5780 & 0.6334 \tabularnewline
    \tabularnewline-
    Efficient-Net B4 & \makecell{Learning Rate : 1e-3 \\ Loss : Cross-Entropy 
    } & 0.7630 & 0.9199 & 0.4114 & 0.5739 & 0.6469 \tabularnewline
    \tabularnewline
    Efficient-Net B4 & \makecell{Learning Rate : CLR  \\ Sampler : $\propto 1/n_j$ \\ Loss : LDAM+DRW \\  }
    & 0.7196 & 0.7867 & 0.5900 & 0.5648 & 0.7699 \tabularnewline
    \tabularnewline
    \bottomrule
\end{tabular}
\end{adjustbox}
\caption{LDAM+Sampling comparison, '4' significantly improves rare-class recall while maintaining decent ValAcc. (DRW refers to \textbf{Deferred Re-Weighting Routine})}
\label{table2}
\end{table}

Deep learning networks for all their might still fare very poorly on highly imbalanced datasets.
Re-sampling and Re-weighting are the most common techniques used to cope with class imbalance problem.

\begin{enumerate}
\item Re-sampling : 
\begin{enumerate}
    \item Oversampling  \cite{relay2016, Zhong2016, Buda2018ASS, Byrd2019WhatIT} : Augmenting the dataset with multiple copies of minority class samples, however since we inherently have low information about the minority class oversampling more often than not leads to overfitting on minority class  \cite{Cui2019ClassBalancedLB}.
    
    \item Undersampling  \cite{He2009LearningFI, Japkowicz2002TheCI, Buda2018ASS}: Undersampling is achieved by rejecting samples from the more-frequent classes. Since we are loosing out on purpose in order to equalize the class-count, undersampling technique aren't possible in case of high class imbalance  \cite{Cui2019ClassBalancedLB}.
\end{enumerate}

\item Re-weighting  \cite{Huang2016LearningDR, Huang2019DeepIL}: Different set of weights ($\propto 1/n_j$,where $n_j$ = total samples of $j^{th}$ class) are assigned to different classes. However re-weighting techniques cause instability in network's optimization under extreme class imbalance  \cite{Cui2019ClassBalancedLB, relay2016, Byrd2019WhatIT}.
\end{enumerate}

Both re-sampling and re-weighting conclusively aim to augment the training distribution to become much more identical to the test distribution. However, due to the aforementioned flaws performance of minority class is generally increased on the cost of the network's ability to learn the majority class well.

Cao et al.  \cite{LDAM} designed a label-distribution aware loss function (LDAM) that regularizes the minority class much more strongly than the majority class, motivating the network to improve generalization on the minority class without suppressing the network’s ability to learn the majority class. Strong regularisation here can be understood in terms of enforcing bigger margins for the minority class as compared to the majority class. Moreover this approach is orthogonal to re-weighting and re-sampling, ensuring flexibility depending on level of imbalance in one's dataset.

In the same work, Cao et al.  \cite{LDAM} proposed a “deferred re-balancing training" procedure which divides the training procedure into two stages. The first stage uses Empirical Risk Minimization with LDAM loss, learning a good initial representation. The second stage employs re-weighted LDAM loss with a smaller learning rate. The main rationale behind this is to bypass the problems caused by re-weighting in the optimization process of a Neural Network by first learning a good initial representation and then optimizing on that. We also employ a re-sampling scheme ($ \propto 1/n_j  $) orthogonal to the LDAM+DRW routine.
\autoref{table2} presents our results with LDAM.

In the next subsection on CLR, we briefly discuss the advantages of Cyclic Learning Rate and later establish it's compatibility and usefulness for imbalance scenarios.

\subsection{Cyclical Learning Rates (CLR)} \label{clr}

\begin{table}[b]
\centering
\begin{tabular}{|l|l|}
\hline
\textbf{Training Details} & \textbf{Values} \\ \hline
Upper Bound & 1e-3 \\ \hline
Lower Bound & 1e-5 \\ \hline
Stepsize & 2 \\ \hline
Functional Form & Triangular \\ \hline
\end{tabular}
\caption{Training Details for CLR setup}
\label{CLRtable}
\end{table}

Learning rate is responsible for scaling the gradients at each weight update and is one of the most important hyper-parameters to tune while training a deep neural network as too small a learning rate will encourage very small steps and hence the network might not converge at all, whereas too high a learning rate will propel divergent behavior. The optimal learning rate depends on the network’s loss surface and usually is not feasible to calculate.

The cyclical learning rate  \cite{CyclicalLR} oscillates between a range of values, going against the conventional wisdom of exponentially/step-wise decreasing the learning rate as training progresses. 
The advantages of doing this are - 
\begin{enumerate}
  \item Stuck on a sharp minimum  \cite{Li2018VisualizingTL} - Networks with flatter minima tend to be more robust than the ones with sharp minima, as flatter minima ensure that we are in optimal minima region in the test loss surface as well and hence generalize better, periodically increasing the value of learning rate will help to get out of the sharp minima more quickly.
  \item Stuck on saddle points  \cite{Jin2017HowTE, CyclicalLR} -  When training a Deep network it is very likely that the loss surface topology contains a lot of saddle points. Thus having per periodic boost of high learning rate is very useful as it helps in traversing the saddle points more quickly (since the gradient value is already very low here).
\end{enumerate}

Experimental values are given in \autoref{CLRtable}.\\
In combination with CLR, we use Stochastic Weight Averaging (SWA) which is a very promising regularization technique. We outline it's benefits for our problem and the setup details in the following subsection.

\subsection{Stochastic Weight Averaging (SWA)} \label{swavg}

\begin{table}[b]
\begin{adjustbox}{width=\columnwidth,center}
\begin{tabular}{cccccccccc}
    \toprule
    \multirow{2}{*}[-4pt]{\thead{Model}} &
    \multirow{2}{*}[-4pt]{\thead{Training Details}} &
    \multirow{2}{*}[-4pt]{\thead{Validation Accuracy}} &
    
    \multicolumn{4}{c}{\thead{Recall}}\\
    \cmidrule(lr){4-7}
    &&& 0 & 1 & 2 & 3\\
    \midrule
    Efficient-Net B4 & \makecell{Learning Rate : 1e-3 \\ Loss : Cross-Entropy} & 0.7630 & 0.9199 & 0.4114 & 0.5739 & 0.6469 \tabularnewline
    \tabularnewline
    
    Efficient-Net B4 & \makecell{Stochastic Weight Averaging : No  \\Learning Rate : CLR  \\ Sampler : $\propto 1/n_j$ \\ Loss : LDAM+DRW \\}
    & 0.7196 & 0.7867 & 0.5900 & 0.5648 & 0.7699 \tabularnewline
   \tabularnewline
   
    Efficient-Net B4 & \makecell{Stochastic Weight Averaging : Yes \\ Learning Rate : CLR  \\ Sampler : $\propto 1/n_j$ \\ Loss : LDAM+DRW \\}
    & 0.7292 & 0.8098 & 0.6017 & 0.5409 & 0.7436 \tabularnewline
    \tabularnewline
    \bottomrule
\end{tabular}
\end{adjustbox}
\caption{SWA Experiment}
\label{SWA}
\end{table}

Another go-to methodology machine learning practitioners generally adopt while training models is ensemble learning. Ensemble learning improves predictions by combining [for example voting, averaging etc] results of various models. However when training Deep Neural Networks it is not possible to train multiple models on the dataset due to time and compute constraints.

Garipov et al.  \cite{Garipov2018LossSM} in their work on Fast Geometric Ensembles showed that using cyclical learning rates with stochastic gradient descent traversed on the periphery of the optimal weights but never quite reached it’s center, They selected the network with weights on the periphery to form the ensemble. This helped in training the ensemble in the time required to train one network. 

Stochastic Weight Averaging  \cite{Izmailov2018AveragingWL} uses the same setup i.e. high frequency cyclical/constant learning rate with SGD to traverse around the optimal weight set, and then does averaging in the weight domain only at different snapshots of training. This allows weights to reach the much desired optimal set. The advantages of this are following:

\begin{enumerate}
  \item Faster inference time compared to Garipov et al.  \cite{Garipov2018LossSM},  as we only have one model as the end result, compared to waiting for k results from k models.
  \item Given that the underlying data distribution is the same, it is fair to assume that the test and train datasets will have similar loss surfaces. Thus it makes much more sense to aim for a more flatter minima while training than a sharp one [even if it leads to higher training error], as it will ensure that we are in an optimal minima region in the test loss surface as well, leading to a more robust network. 
\end{enumerate}

We find that using SWA in combination with Adam optimizer and the CLR setup we were able to significantly improve the low/mid class accuracy and subsequently train a more robust network \autoref{SWA}.

In the next section we present our brief insights connecting remote sensing knowledge with research in machine learning. The bands are a key component and must be studied in more detail for better cross-linking when being used with neural networks.

\section{Training On Subset of Bands} \label{msbands}

\begin{table}[b]
\begin{adjustbox}{width=\columnwidth,center}
\begin{tabular}{cccccccccc}
    \toprule
    \multirow{2}{*}[-4pt]{\thead{Model}} &
    \multirow{2}{*}[-4pt]{\thead{Training Details}} &
    \multirow{2}{*}[-4pt]{\thead{Validation Accuracy}} &
    
    \multicolumn{4}{c}{\thead{Recall}}\\
    \cmidrule(lr){4-7}
    &&& 0 & 1 & 2 & 3\\
    \midrule
    Efficient-Net B4 & \makecell{Bands: All \\ Stochastic Weight Averaging : Yes \\ Learning Rate : CLR  \\ Sampler : $\propto 1/n_j$ \\ Loss : LDAM+DRW \\}
    & 0.7292 & 0.8098 & 0.6017 & 0.5409 & 0.7436 \tabularnewline
    \tabularnewline
    Efficient-Net B4 & \makecell{Bands: 4,3,2 \\ Stochastic Weight Averaging : Yes \\ Learning Rate : CLR  \\ Sampler : $\propto 1/n_j$ \\ Loss : LDAM+DRW \\}
    & 0.7047 & 0.7996 & 0.5356 & 0.5475 & 0.6428 \tabularnewline
    \tabularnewline
    Efficient-Net B4 & \makecell{Bands: 5,4,3 \\ Stochastic Weight Averaging : Yes \\ Learning Rate : CLR  \\ Sampler : $\propto 1/n_j$ \\ Loss : LDAM+DRW \\}
    & 0.7022 & 0.7792 & 0.5719 & 0.5780 & 0.6334 \tabularnewline
    \tabularnewline
    Efficient-Net B4 & \makecell{Bands: 6,5,2 \\ Stochastic Weight Averaging : Yes \\ Learning Rate : CLR  \\ Sampler : $\propto 1/n_j$ \\ Loss : LDAM+DRW \\}
    & 0.7441 & 0.8156 & 0.5941 & 0.6138 & 0.7584 \tabularnewline
    \tabularnewline

    \bottomrule
\end{tabular}
\end{adjustbox}
\caption{Using subset of bands}
\label{bands}
\end{table}

Multi-spectral Images (MSI) are described by 3 to 10  narrow spectral bands. This high spectral information is very beneficial as by combining different spectral bands we can infer different information, leading up to terabytes of data produced per day.

Since adjacent bands in MSI are highly correlated, there is a lot of redundancy in our data. This contrary to conventional wisdom, leads to degradation of accuracy on increasing the number of bands in MS images  \cite{BS}, also using too many spectral bands incur high computational cost as well as more inference time. 

Thus it makes sense to use only those spectral bands which motivate the network to learn better feature representations for separating specific classes. The selected band performance is often conditional on many aspects of the classification pipeline such as  the nature of the adopted classifier and its parameter configurations  \cite{HBS}. 

A major hurdle was deciding the importance of each spectral band, since there is not a lot of literature specific to neural networks. We experimented with three different band combinations based on their characteristics.  \cite{Focareta2015}

\begin{enumerate}

\item 4-3-2: \textbf{Natural Color} ~ This band combination results in the image appearing as perceived by the human eye.
\item 5-4-3: \textbf{Near Infrared Composite} ~ This combination contains near-infrared(5), red(4), green(3) bands, This combination is particularly useful while analyzing vegetation, crops and wetlands as it is able to capture the near-infrared light reflected by chlorophyll.
\item 6-5-2: \textbf{Agriculture} ~ It is a combination of SWIR-1 (6), near-infrared (5) and blue (2). The short-wave and near infrared allows this combination to be used for crop monitoring.
\end{enumerate}

As observed in \autoref{bands}, the combination 6-5-2 seems to work the best for the given dataset.

We believe the original dataset is small but inherently complex due to overlap of several spectral bands and thus data augmentation is very beneficial. The next section expands on the data generation component of our project.

\section{Generating Synthetic Images} \label{gans}

\begin{table}[b]
\begin{adjustbox}{width=\columnwidth,center}
\begin{tabular}{cccccccccc}
    \toprule
    \multirow{2}{*}[-4pt]{\thead{Model}} &
    \multirow{2}{*}[-4pt]{\thead{Training Details}} &
    \multirow{2}{*}[-4pt]{\thead{Validation Accuracy}} &
    \multicolumn{4}{c}{\thead{Recall}}\\
    \cmidrule(lr){4-7}
    &&& 0 & 1 & 2 & 3\\
    \midrule
    \tabularnewline 
    
    Efficient-Net B4 & \makecell{Learning Rate : 1e-3 \\ Loss : Cross-Entropy \\ Dataset : Original } & 0.76 & 0.9199 & 0.4114 & 0.5739 & 0.6469 
    \tabularnewline
    \tabularnewline 
    
    Efficient-Net B4 & \makecell{Stochastic Weight Averaging : No  \\Learning Rate : 1e-3  \\ Sampler : $\propto 1/n_j$ \\ Loss : Cross-Entropy \\ Dataset : GAN-Augmented}
    & 0.67 & 0.6815 & 0.6619 & 0.6258 & 0.7521 \tabularnewline
    \tabularnewline
    
    Efficient-Net B4 & \makecell{Bands: 6,5,2 \\ Stochastic Weight Averaging : Yes \\ Learning Rate : CLR  \\ Sampler : $\propto 1/n_j$ \\ Loss : LDAM+DRW \\ 
    Dataset : Original}
    & 0.74 & 0.8156 & 0.5941 & 0.6138 & 0.7584 \tabularnewline
    \tabularnewline
    
    Efficient-Net B4 & \makecell{Stochastic Weight Averaging : Yes \\ Learning Rate : CLR  \\ Sampler : $\propto 1/n_j$ \\ Loss : LDAM+DRW \\
    Dataset : GAN-Augmented}
    & 0.70 & 0.7459 & 0.5842 & 0.5540 & 0.7436 \tabularnewline
    \tabularnewline
    \bottomrule
\end{tabular}
\end{adjustbox}
\caption{GAN Augmented Dataset Comparison}
\label{GANtb}
\end{table}

%%%%%%%%%%%%%%%%%%%%%%%%%%%
\begin{table}[htbp]
\centering
\begin{tabular}{|l|l|}
\hline
\textbf{Training Details} & \textbf{Values} \\ \hline
Resolution & 64x64 \\ \hline
Epochs & 45 \\ \hline
Learning Rate & 2e-4 \\ \hline
$\beta1$ & 0.5 \\ \hline
$\beta2$ & 0.999 \\ \hline

\end{tabular}
\caption{Training Details for GAN}
\label{GANtraintable}
\end{table}
The introduction of Generative Adversarial Networks (GANs)  \cite{Goodfellow2014GenerativeAN} sprung up many exciting research directions, the field has grown steadily with numerous applications in image super-resolution, in-painting, image-to-image translation, image enhancement (For example, earth observation/remote sensing   \cite{Liu2018PSGANag, Tsagkatakis2019SurveyOD}).

Standard data augmentation has been used as a go-to technique for enhancing generalizability. Generative adversarial networks offer a novel method for data augmentation  \cite{Sandfort2019DataAU}, but have still not been adopted by either the earth observation or remote sensing community.
% AGAN consists of two models (the Generator and the Discriminator), with both having adversarial optimization objectives. The generator’s goal is to produce samples that are indistinguishable from the real samples for the discriminator, whereas the discriminator’s objective is to discriminate the real samples from the generated ones. This is a mini-max optimization problem and the desired solution turns out to be the Nash equilibrium where the generator has become so good that the discriminator can no longer identify what’s real and what’s fake to a reasonable accuracy i.e.  it is as good as taking a random guess   50 percent accuracy
We use DC-GAN  \cite{Radford2015UnsupervisedRL}, which employs deep convolutional neural networks for both the Generator (G) and Discriminator (D), to generate synthetic images for the low represented classes as a form of data-augmentation to equalize the number of samples of each class. We only operate on a subset of bands (6-5-2), since it is easier to critic the visual perceptibly of images this way than all the bands combined.

The main motivation behind equalizing the number of classes was to make the network learn improved discriminatory features and hence becomes more robust. 

\begin{figure}[htbp]
  \centering
    \begin{subfigure}[b]{0.49\linewidth}
    \includegraphics[width=\textwidth]{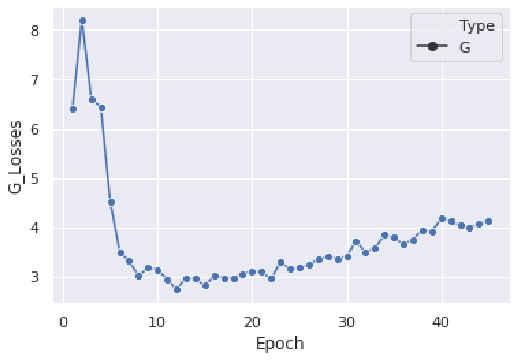}
    \caption{Generator (G) Loss}
  \end{subfigure}
  \begin{subfigure}[b]{0.49\linewidth}
    \includegraphics[width=\textwidth]{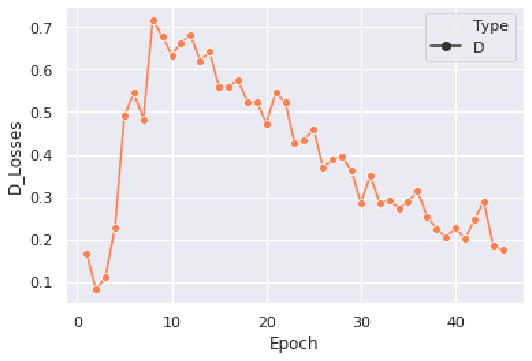}
    \caption{Discriminator (D) Loss}
    \end{subfigure}

\caption{GAN Losses}
\label{fig4}
\end{figure}

\begin{figure*}
	\centering
	\includegraphics[width=1\linewidth, keepaspectratio]{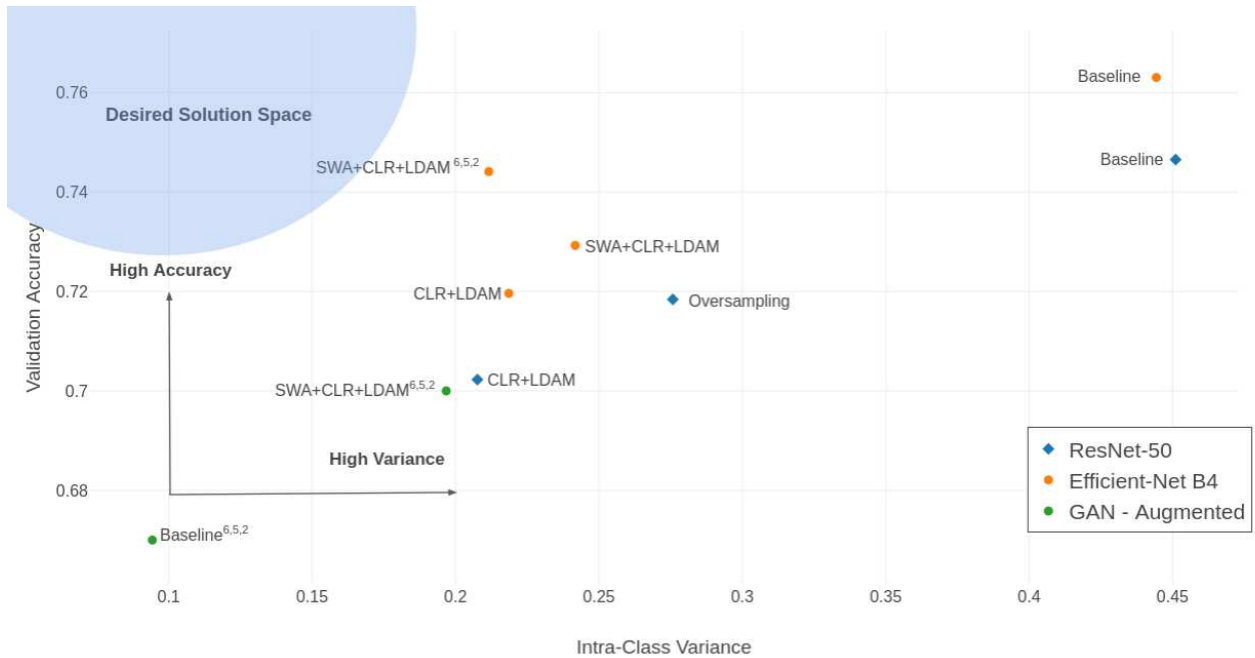}
	\caption{Plot of various training methodologies w.r.t Validation accuracy and Intra-class variance \\ 
    The solutions at the top-left section (more accurate, less variant respectively) of the graph are most desirable (i.e. robust).}
	\label{fig1}
\end{figure*}

We monitored the visual perceptibly of generated images over the training period (60 epochs) and found that the network converges at about 45 epochs, see \autoref{fig3}.
The final dataset (GAN-Augmented) consisted of 120,000 images with 30,000 images from each class. The architectures used for D and G are kept same as described in  \cite{Radford2015UnsupervisedRL}. Training details are shown in \autoref{GANtraintable} and Loss plots are in \autoref{fig4}.

\autoref{GANtb} demonstrates the results we obtained from the network (in combination with various training methodologies) on the GAN-Augmented dataset, a significant increase in the per-class accuracies of rare-classes was observed.

\section{Results}

\begin{table}[htbp]
\begin{adjustbox}{width=\columnwidth,center}
\begin{tabular}{cccccccccc}

    \toprule
    \multirow{2}{*}[-4pt]{\thead{Model}} &
    \multirow{2}{*}[-4pt]{\thead{Training Details}} &
    \multirow{2}{*}[-4pt]{\thead{Validation Acc}} &
    \multirow{2}{*}[-4pt]{\thead{Balanced Validation Acc}} &
    \multirow{2}{*}[-4pt]{\thead{Intra-Class Variance}} 
    \tabularnewline
    \tabularnewline
    \tabularnewline
    \midrule
    ResNet-50 & \makecell{Learning Rate : 1e-3  \\ Loss : Cross-Entropy} & 0.7465 & 0.6123 & 0.4510 \tabularnewline
    \tabularnewline
    ResNet-50 & \makecell{Learning Rate : 1e-3  \\ Sampler : $\propto 1/n_j$ \\ Loss : Cross-Entropy \\  }
    & 0.7184 &  0.6467 & 0.2757 \tabularnewline
    \tabularnewline
    ResNet-50 & \makecell{Learning Rate : CLR  \\ Sampler : $\propto 1/n_j$ \\ Loss : LDAM+DRW \\  }
    & 0.7022 & 0.2076 & 0.6406 \tabularnewline
    \tabularnewline
    Efficient-Net B4 & \makecell{Learning Rate : 1e-3 \\ Loss : Cross-Entropy} & 0.7630 & 0.6380 & 0.4443 \tabularnewline
    \tabularnewline
        
    Efficient-Net B4 & \makecell{Learning Rate : CLR  \\ Sampler : $\propto 1/n_j$ \\ Loss : LDAM+DRW \\  }
    & 0.7196 & 0.6779 & 0.2185 \tabularnewline
    \tabularnewline
    Efficient-Net B4 & \makecell{Stochastic Weight Averaging : Yes \\ Learning Rate : CLR  \\ Sampler : $\propto 1/n_j$ \\ Loss : LDAM+DRW \\}
    & 0.7292 & 0.6740 & 0.2417 \tabularnewline
    \tabularnewline
    Efficient-Net B4 & \makecell{Bands: 6,5,2 \\ Stochastic Weight Averaging : Yes \\ Learning Rate : CLR  \\ Sampler : $\propto 1/n_j$ \\ Loss : LDAM+DRW \\}
    & 0.7441 & 0.6955 & 0.2115 \tabularnewline
    \tabularnewline
    Efficient-Net B4 & \makecell{Stochastic Weight Averaging : No  \\Learning Rate : 1e-3  \\ Sampler : $\propto 1/n_j$ \\ Loss : Cross-Entropy \\ Dataset : GAN-Augmented}
    & 0.67 & 0.6803 & 0.0942 \tabularnewline
    \tabularnewline
    Efficient-Net B4 & \makecell{Stochastic Weight Averaging : Yes \\ Learning Rate : CLR  \\ Sampler : $\propto 1/n_j$ \\ Loss : LDAM+DRW \\ Dataset : GAN-Augmented}
    & 0.70 & 0.6569 & 0.1967 \tabularnewline
    \tabularnewline
    
    \bottomrule
\end{tabular}
\end{adjustbox}
\caption{Performance Metrics Comparison}
\label{MetricsTable}
\end{table}

\begin{figure*}
	\centering
	\includegraphics[width=1\linewidth, keepaspectratio]{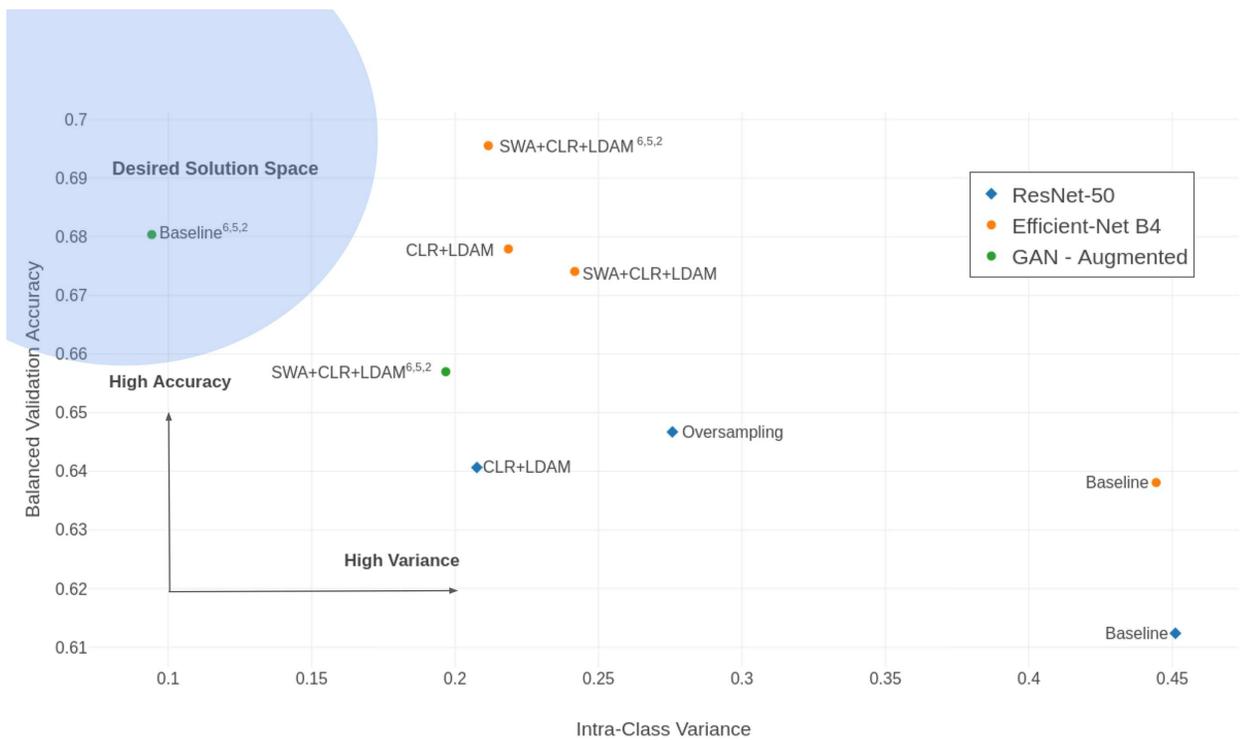}
	\caption{Plot of various training methodologies w.r.t Balanced-Validation accuracy and Intra-class variance\\ 
	We observe various models which were performing very well as per the \textbf{vanilla validation-accuracy} plummet when plotted w.r.t. \textbf{balanced validation accuracy} thus exposing the deep-rooted focus on the frequent class and futility of ValAcc.
	}
	\label{fig2}
\end{figure*}

\begin{figure*}[hbtp]
  \centering
    \begin{subfigure}[b]{0.45\linewidth}
    \includegraphics[width=\textwidth]{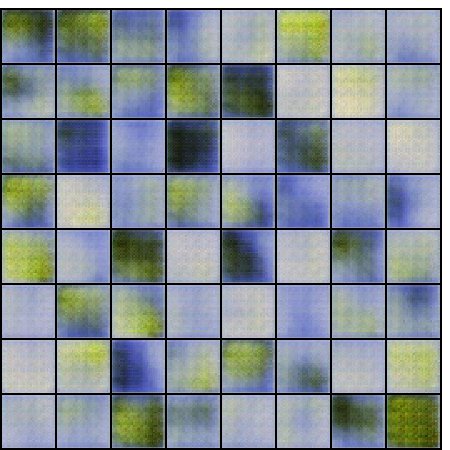}
    \caption{Initial: 7 epochs}
  \end{subfigure}
  \begin{subfigure}[b]{0.45\linewidth}
    \includegraphics[width=\textwidth]{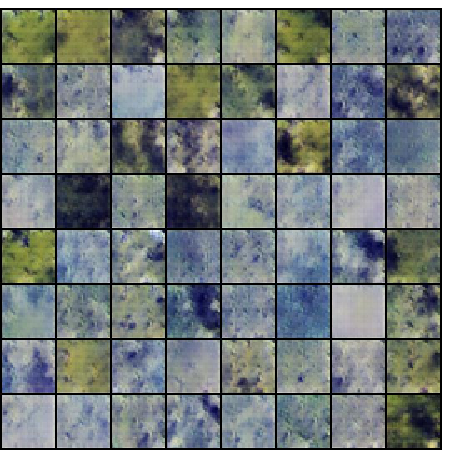}
    \caption{Midway: 20 epochs}
  \end{subfigure}
  \hfill
  \begin{subfigure}[b]{0.45\linewidth}
    \includegraphics[width=\textwidth]{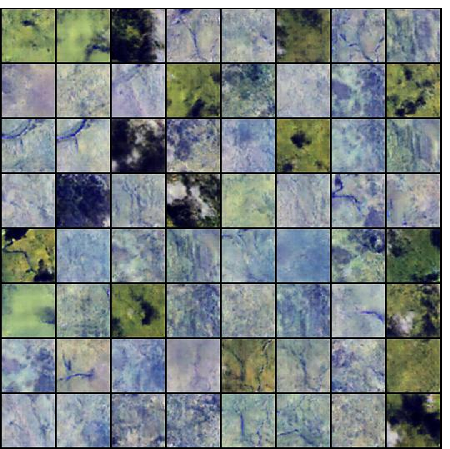}
    \caption{Final: 45 epochs}
  \end{subfigure}
    \begin{subfigure}[b]{0.445\linewidth}
  \fbox{\includegraphics[width=\textwidth]{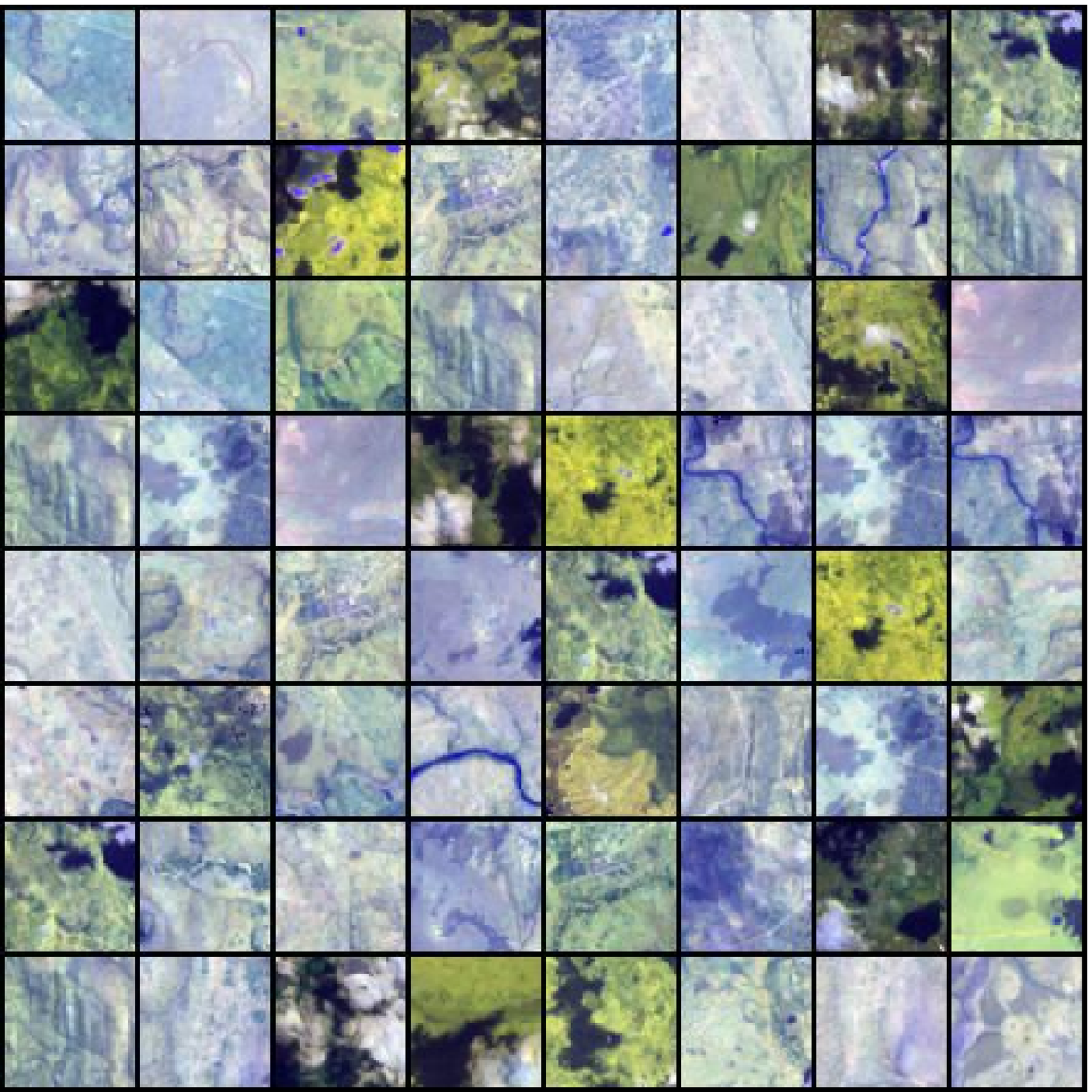}} 
    \caption{Original Dataset c/o W\&B Inc.(ILRI-Cornell-UCSD)  \cite{droughtdataWandB}}
  \end{subfigure}

\caption{Sample images from our GAN training stages}
\label{fig3}
\end{figure*}

We evaluate various techniques in a combination setting to facilitate training of robust Deep Neural Networks. We provide baseline metrics for both architectures (ResNet-50 and Efficient-Net B4) in \autoref{NNarch}, and observe that the baselines fall prey to overfitting owing to high class imbalance. \autoref{table2} advocates LDAM loss as a label-dependent regularizer which leads to a reduction in Intra class variance (ICV) and improvements in balanced-accuracy, see \autoref{fig2}. 
We observe that performance of SWA+LDAM+CLR (all bands, \autoref{MetricsTable} - 6) performs  compared to SWA+LDAM+CLR$^{6,5,2}$ (\autoref{MetricsTable} - 7).

Lastly we present the results of baseline as well as SWA+LDAM+CLR$^{6,5,2}$ on GAN-Augmented dataset. There is a substantial decrease in ICV while maintaining decent balanced-accuracy in the baseline experiment, indicating that the network was able to learn better discriminative features for all rare-classes. The SWA+LDAM+CLR$^{6,5,2}$ though under-performing on aspect of per-class accuracy leads to considerable decrease in ICV, see \autoref{fig2}.

\subsection{Limitations}
    \begin{itemize}
        \item In  \autoref{fig2} (BalAcc vs ICV) we observe one outlier result: \textit{SWA+CLR+LDAM $^{6,5,2}$ -GAN Augmented}, as per our trend this should have been the best result (instead it is the \textit{Baseline - GAN Augmented}). This exception may be attributed to incomplete insights on the GAN data interaction with our network modifications.
        \item Problems with generating data for all spectral bands. There is a lack of empirical data  to ascertain quality of output data in such scenario.  \cite{Liu2018PSGANag},  \cite{Tsagkatakis2019SurveyOD},  \cite{Kerdegari2019SemisupervisedGF}. We expect improvements with higher-diversity images  \cite{Ghosh2018MultiagentDG}.
        \item We did not explore alternative generative models ex. Kernel-based GANs  \cite{Mehrjou2019KernelGuidedTO}, Variational Autoencoders family (VQ-VAEs  \cite{Oord2017NeuralDR-VQVAE, Razavi2019VQVAE2}, hybrid VAE-GAN  \cite{Larsen2015VAEGAN}). 
        \item No class-activation mapping for model explanation or other interpretability mechanism  \cite{Zhou2016LearningDF, Selvaraju2016GradCAMVE, Hooker2019ABF}.
        \item We did not incorporate adversarial training/defense.
    \end{itemize}

\section{Conclusion} \label{end}

There is a lot of focus on handling or curbing the adverse effects of imbalanced data. Mitigating class imbalance is an important research area, as it will allow trust-worthy solutions in the form of deep neural networks in many eclectic fields. As per trend, deep learning networks are tuned to maximize the total accuracy over the entire dataset, thus focusing on the majority-class samples. As a result, the models under-perform on minority class(es) samples leading to bad intra-class generalization and low robustness. 

We provide a comparative overview of diverse yet latest methodologies for operating on skewed datasets that is suffering from class-imbalance problems. This diverse set of techniques ranges from discussions on state-of-the-art convolutional neural network architectures, label-dependent loss functions, learning-rate routines, generating Deep Neural Network ensembles and finally generating data samples using DC-GAN. 

We conclusively aspire to serve as a toolkit for practitioners and researchers suffering from skewed data problems in their respective fields as we present the work to other domain-experts, especially those dealing with multiple minority classes. Since our ensemble methodology doesn't overfit on the rare classes but tries to generalize on the non-major classes thus achieving a trade-off on overall accuracy but high robustness.

\subsection*{Acknowledgements}

We would like to thank \href{https://scholar.google.co.in/citations?user=PPsneeEAAAAJ}{Meenakshi Sarkar} and \href{https://scholar.google.com/citations?user=KpbexaEAAAAJ}{Shivam Saboo} for insightful discussions, also the anonymous reviewers for their valuable feedback on the draft.
Authors would like to give a shout-out to \href{http://wandb.com}{Weights \& Biases} and to the \href{https://www.climatechange.ai/ICLR2020_workshop.html}{ICLR'20 CCAI Workshop}'s Mentorship Program. 

PP extends special thanks to \href{https://scholar.google.com/citations?user=BUfKuTYAAAAJ}{Debasish Ghose (IISc-B)} and \href{http://krikamol.org}{Krikamol Muandet (MPI-IS)} for supporting this work.

{\small                                                        
\bibliographystyle{ieee_fullname}
\bibliography{egbib}
}

\end{document}